# SegmentAnyTree: A sensor and platform agnostic deep learning model for tree segmentation using laser scanning data

Maciej Wielgosz [a][†][*], Stefano Puliti [a][*], Binbin Xiang [b], Konrad Schindler [b], Rasmus Astrup [a]

[a] Norwegian Institute of Bioeconomy Research (NIBIO), 1433 Ås, Norway

[b] Photogrammetry and Remote Sensing, ETH Zürich, 8093 Zürich, Switzerland

† Corresponding author (maciej.wielgosz@nibio.no)

* These authors contributed equally to this study

## Abstract

This study focuses on advancing individual tree crown (ITC) segmentation in lidar data, developing a sensor- and platform-agnostic deep learning model transferable across a spectrum of airborne (ULS), terrestrial (TLS), and mobile (MLS) laser scanning data. In a field where transferability across different data characteristics has been a longstanding challenge, this research marks a step towards versatile, efficient, and comprehensive 3D forest scene analysis.

Central to this study is model performance evaluation based on platform type (ULS vs. MLS) and data density. This involved five distinct scenarios, each integrating different combinations of input training data, including ULS, MLS, and their sparsified versions, to assess the model's adaptability to varying resolutions and efficacy across different canopy layers. The core of the model, inspired by the *PointGroup* architecture, is a 3D convolutional neural network (CNN) with dedicated prediction heads for semantic and instance segmentation. The model underwent comprehensive validation on publicly available, machine learning-ready point cloud datasets. Additional analyses assessed model adaptability to different resolutions and performance across canopy layers.

Our results reveal that point cloud sparsification as an augmentation strategy significantly improves model performance. It extends the model's capabilities to sparse LiDAR data and boosts detection and segmentation quality in dense, complex forest environments. Notably, the model showed consistent performance for point clouds with densities >50 points $m^{-2}$ but exhibited a drop in performance at the sparsest level (10 points $m^{-2}$), mainly due to increased omission rates. Benchmarking against current state-of-the-art methods established the proposed model's superior performance on multiple open benchmark datasets. For example, on the *LAUTx* dataset, our method outperformed *Point2Tree* and *TLS2trees* by ≈20-30% in detection rate, omission rate, commission rate and F1 score. Our experiments also set new performance baselines for the *Wytham Woods* and

*TreeLearn* datasets. The comparison highlights the model's superior segmentation skill, mainly due to better detection and segmentation of understory trees below the canopy, with reduced computational demands compared to other recent methods.

In conclusion, the present study demonstrates that it is indeed feasible to train a sensor-agnostic model that can handle diverse laser scanning data, going beyond current sensor-specific methodologies. Further, our study sets a new baseline for tree segmentation, especially in complex forest structures. By advancing the state-of-the-art in forest lidar analysis, our work also lays the foundation for future innovations in ecological modeling and forest management.

# 1. Introduction

Obtaining information on individual trees to support smaller-scale and multifunctional forest management has been a central area of research for the last three decades. A wide variety of approaches, often referred to as individual tree crown (ITC) methods, has been proposed to segment individual trees in lidar data. We presently witness a renaissance of research into single tree methods, fueled by increased access to high-resolution point clouds thanks to sensor improvements, combined with progress in deep learning. Most existing ITC methods are tailored to a specific type of lidar data as input, with distinct approaches and traditions for airborne lidar (ALS) or for proximally sensed terrestrial (TLS), mobile (MLS), or drone (ULS) lidar. Comparatively, little effort went into developing platform- and sensor-agnostic methods or models that are transferable to any new lidar data without carefully re-tuning various key parameters and settings. Such an agnostic ITC method would offer unprecedented versatility during deployment, as it could be applied to any given lidar data and deliver consistent predictions across datasets. Once available, such models would likely also boost the operational use of ITC methods by eliminating the need for forest practitioners and researchers to navigate the vast landscape of ITC methods and find the most suitable one for their specific lidar datasets.

Since the inception of ITC approaches for airborne laser scanning (ALS) data (Hyyppa et al. 2001), the state-of-the-art (SOTA) for tree detection and segmentation algorithms has improved only marginally. An evaluation of two decennially spaced ITC benchmark efforts—one by Kaartinen et al. (2012) and the other by Cao et al. (2023)—indicates only slight enhancements in metrics such as detection rates or F1 scores. In both studies, the main limitation was found to be the low detection rates (52% on average Kaartinen et al. (2012)), driven by the poor detection rates of co-dominant trees (i.e. trees in the dominant canopy but very close to each other) and the severe under-detection of suppressed or understory trees (i.e., trees entirely covered by the main canopy and not visible from above). The relatively large omission rates for smaller trees resulted in a scarce uptake in operational forest planning. One of the main drivers of the poor detection of co-dominant and dominated trees is the reliance on canopy height models (CHMs), which reduce the scan points to a top-down view (canopy top height above ground) and, by definition, miss the understory component. It is evident that advanced ITC methods should operate on raw point clouds in order to utilize the rich information captured along the vertical profile of the forest canopy. Moreover, ITC methods typically start with selecting seed points to initialize tree crown segments. This initialization is critical, as mistakes will irrevocably propagate to all downstream analyses based on those segments. The seed points are often associated with treetops identified in CHMs, using a moving filter with a predefined window size. The window size is a hyperparameter closely linked to the horizontal forest structure, and while attempts have been made to adapt it dynamically based on ALS data properties (e.g., Popescu et al. (2002)), empirical fine-tuning of the window size remains a necessary step for new datasets, limiting transferability.

In parallel to the development of ALS ITC methods, we have, in the past ten years, also witnessed an active development of methods to segment individual trees in very detailed 3D point clouds (e.g., Tao et al. (2015) and Burt et al. (2019)). According to the benchmarking effort by Liang et al. (2018), TLS generally outperforms ALS for ITC. Even so, crown segmentation results are often manually edited, making them a main bottleneck when it comes to unlocking the wealth of information available in proximally sensed lidar data either from the ground (TLS and MLS) or from drones (ULS).

In the past couple of years, driven by rapid advances in AI research and by the availability of open, analysis-ready point cloud benchmark datasets (e.g. Puliti et al. 2023a), significant progress has been made in ITC segmentation (Hakula et al. 2023; Henrich et al. 2023; Straker et al. 2023; Wielgosz et al. 2023; Xiang et al. 2023a). Given enough training data, these models are entirely data-driven, eliminating the need for hyperparameter tuning and manual configuration for different forest structures. Moreover, they exhibit impressive transferability, as demonstrated, for instance, by Wilkes et al. (2023), who successfully applied a model developed based on a small set of plots in an Australian forest (Krisanski et al. (2021)) to a wide variety of forest types worldwide. Recent studies showed that deep-learning methods tend to outperform traditional CHM-based approaches. The larger detection rates can partly be attributed to better detection of understory trees (Xiang et al. 2023b).

Most recent research into deep learning-based ITC segmentation employed very dense TLS, MLS, or ULS point clouds. The level of detail of such point clouds allows for a clear visual distinction between individual trees, enabling manual annotation of suitable training and evaluation data and SOTA accuracy of the resulting ITC segmentation. While advancing the SOTA for terrestrial data, these methods are often custom-tailored to the characteristics of terrestrial data, such as very high point density on stems and low vegetation (e.g. Wilkes et al. 2023). As such, they are, by design, not directly transferrable to airborne laser scanning data collected above the canopy.

Fewer examples exist in the literature for deep learning-based ITC from ALS data, primarily using very dense ALS or ULS data. Windrim and Bryson (2020) showed the first promising results of a deep learning model for ITC segmentation. Straker et al. (2023), proposed the use of the YOLOv5 model for ITC crown segmentation and found that, on the FOR-instance data (Puliti et al. 2023a), it outperformed the commonly used Voronoi segmentation (detection rate= 30.7%), while also not requiring any prior knowledge on the spatial distribution of trees. Xiang et al. (2023b) proposed a more advanced 3D point cloud deep learning model that performs full panoptic (i.e., semantic and instance) segmentation of ULS forest scenes. That model outperformed the one of Straker et al. (2023) by nearly 13% points (detection rate 82.3% vs. 69.6%) and currently constitutes the SOTA on the FOR-instance data. The improvements are, however, limited to very dense ULS data (>1000 points/m$^2$). To broaden the scope and operational impact of ITC inventories, it is necessary to ensure applicability to sparser ALS data (< 500 points/m$^2$). One of the main challenges when using deep learning on ALS data is the difficulty of generating suitable annotated training and validation point clouds since it is more complicated for human operators to separate individual trees in ALS point clouds than TLS/MLS data.

One possible solution is to (approximately) simulate ALS data by synthetically down-sampling proximal laser scans that have been labeled. A central hypothesis of the present article is that in this way, one can extend methods developed for TLS/MLS to handle lower-density ALS data, thus obtaining a new generation of transferable models across various types of input data. The objective of our study was to train and validate a sensor-agnostic model for the segmentation of individual trees in laser scans, applicable to both airborne and terrestrial scans. We show that this can be indeed achieved via synthetic sparsification, augmenting the training set with synthetically sub-sampled point clouds.

## 2. Materials

The study was conducted on a diverse collection of in-house and publicly available forest laser scanning scenes where unique tree identifiers have been manually annotated. The following sections describe in detail the used datasets.

### 2.2 Drone laser scanning data

The openly available FOR-instance benchmark dataset (Puliti et al. 2023b) was used as the source of dense airborne laser scanning data. That benchmark comprises fully annotated ULS point clouds collected over five sites in Norway, Czech Republic, Austria, Australia, and New Zealand. The annotations include per-point unique tree identifiers as well as semantic labels. For this study, the semantic labels were flattened to a binary classification between tree (stems, branches, and leaves) and non-tree (ground, low vegetation) points. To enable meaningful comparisons, the dataset prescribes a fixed split into development data for training and model validation (70% of the area) and test data (30% of the area).

### 2.3 Mobile laser scanning data

As ground-based laser scans, we used the MLS data from Wielgosz et al. (2023), consisting of point clouds collected with a GeoSLAM ZEB-HORIZON (GeoSLAM 2020) at 16 circular plots (400 m$^2$). As for the FOR-instance data, the point clouds are annotated with instance IDs and semantic labels, which we again flatten into a binary tree (stems and crowns) vs. non-tree (ground, coarse woody debris, low vegetation) classification. To enable a direct comparison against the instance segmentation method proposed in Wielgosz et al. (2023), we use the same data split, with 25% of the area in each plot set aside for testing.

### 2.4 Sparsified data

To obtain data from the same sites and recording times but with characteristics similar to ALS data, we synthetically sparsify the ULS and MLS point clouds. Given the complexity and computational needs to simulate long-range LiDAR-based only on short-range LiDAR points, we take the most straightforward approach and randomly subsample the point clouds to point densities of 1000, 500, 100 and 10 points/m$^2$. That range of densities covers current airborne capture scenarios from helicopter-based high-density ALS (500 - 1000 points/m$^2$) through conventional, airplane-based high-density ALS (up to 100 pts/m$^2$) to traditional standard ALS (ca. 10 pts /m$^2$). Figure 1 illustrates the input data sources as well as the sparsified versions.

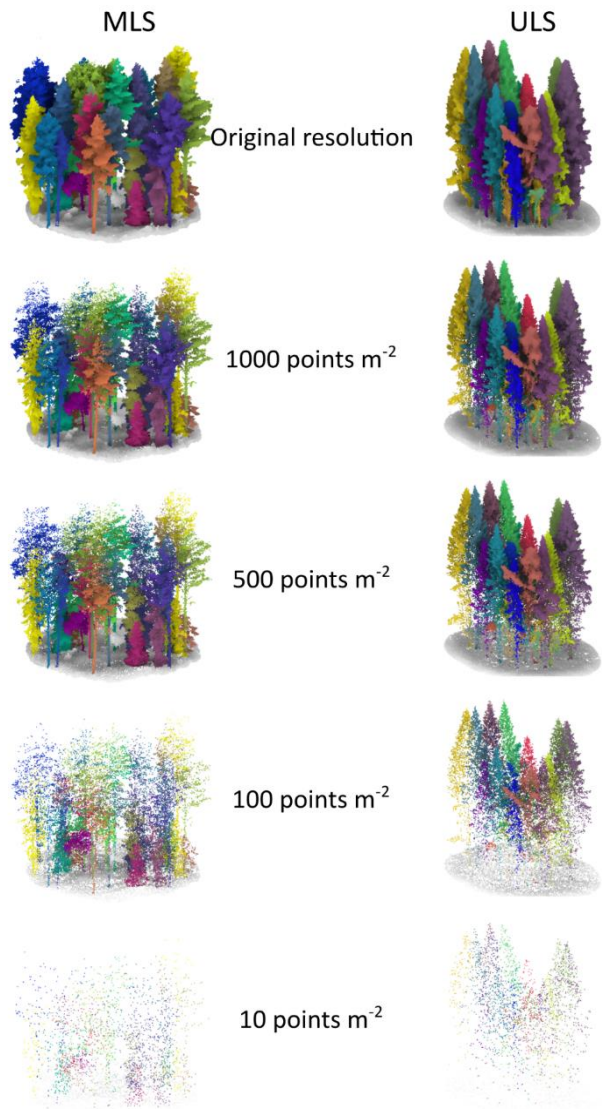

**Figure 1**. Example of one plot from the MLS and one from the ULS data, including the sparsified version of each of the datasets.

## 2.5 Test data

To evaluate the proposed methods, we compiled a comprehensive test dataset comprising the test portions of the above-mentioned ULS (Puliti et al. 2023b) and MLS data and further publicly available, annotated MLS and TLS datasets. Listed by decreasing point density, they are:

- **Wytham Woods dataset** (Calders et al. 2022): This dataset contains data from a single 1.4 ha area with 835 individual trees in Wytham Woods, Oxford, UK. It includes a temperate mixed deciduous forest primarily consisting of Fraxinus excelsior, Acer Pseudoplatanus, and Corylus avellana. The TLS data, collected using a RIEGL VZ-400 (RIEGL Laser Measurement Systems GmbH) in leaf-off conditions, were captured from scan stations on a 20 m x 20 m grid. This dataset is representative of scenarios with high data quality and of structurally diverse broadleaved forests.

- **TreeLearn test dataset** (Henrich et al. 2023): comprising a single area with 156 trees, this dataset has been captured by MLS (GeoSLAM ZEB-HORIZON) in leaf-off conditions. Dominated by mature Beech forest, it represents a very common forest type in Europe.

Henrich et al. (2023) reported tree instance segmentation with an F1-score of 98% for this area.

- **LAUTx dataset** (Tockner et al. 2022): this dataset consists of six inventory plots with corresponding MLS data (captured with GeoSLAM ZEB-HORIZON), featuring a variety of forest types (broadleaved, coniferous, and mixed) and forest structures (single or dual-layered). Wielgosz et al. (2023) used it to establish a baseline for tree instance segmentation and achieved a detection rate of 0.55 and an F1-score of 0.67.

- **NIBIO MLS test data**: MLS data was collected using a GeoSLAM ZEB-HORIZON for 16 circular field plots of 400 m2 area in a boreal forest, which is actively managed for timber production. These data formed the basis for developing the Point2Tree method. Further details on this datset can be found in Wielgosz et al. (2023).

- **FOR-instance test data**: The designated testing portion from the FOR-instance dataset, described above in the context of drone-based laser scanning. To underpin the performance across different forest types, the evaluation was done separately for subsets collected in different regions, named NIBIO, CULS, TU_WIEN, RMIT, and SCION. To date, the studies by Straker et al. (2023) and Xiang et al. (2023a) used the FOR-instance data and provided baseline values for segmentation.

- **Sparsified FOR-instance test data**: For each of the FOR-instance test point clouds we included sparsified versions created by random subsampling, as described above in Section 2.4. They serve to assess performance across different data densities.

## 3. Methods

In this study, to assess the effectiveness of developing sensor- and platform-agnostic models, we adopted an approach whereby the model form and hyperparameters were kept constant while we modified the input training data. As such we trained several models using either ULS or MLS data and different combinations of these and their respective sparsified versions. The models were benchmarked against a comprehensive selection of open datasets. Figure 2 provides an overview of the adopted workflow.

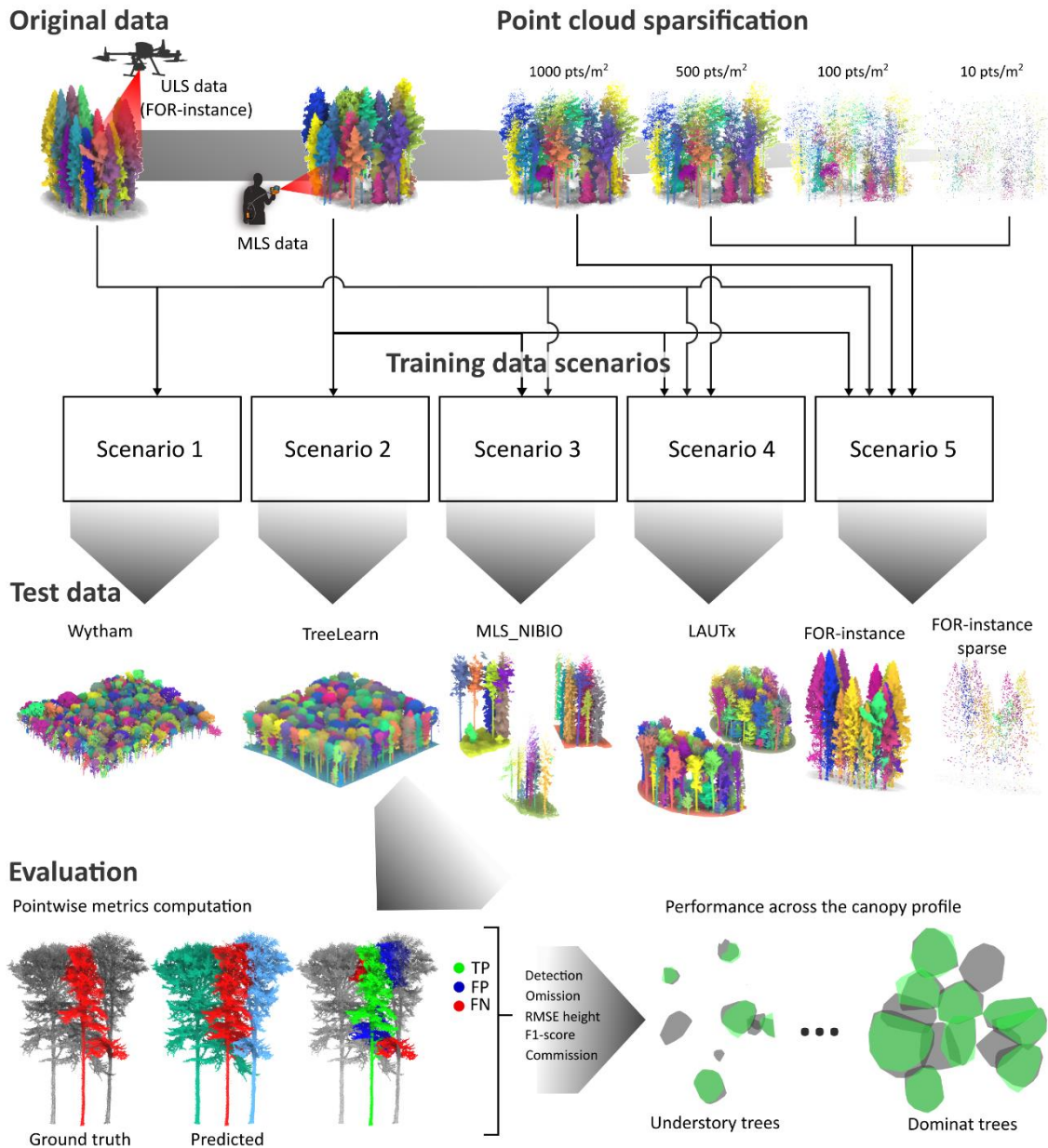

**Figure 2**. Schematic visualization of the implemented workflow for the training and evaluation of the proposed models.

### 3.1 Testing different input data and augmentation scenarios

To assess the performance of the different models based on the input of different point cloud data collected either through terrestrial platforms (MLS) data or through airborne ones (ULS) and at different resolutions, we evaluated the following two aspects:

- **Platform**: here, we compared the performance of the model trained on only ULS data against the one trained on MLS data alone. This allows to better understand the versatility of models trained on either of these data sources regarding their transferability to a broad variety of laser scanning datasets. Further, this provides an understanding of the relative contributions of the different raw data sources to the creation of fully agnostic models.

- **Density**: here, we compared the effect of including sparsified versions of the data in the training of the model as a form of augmentation with the aim to extend the model's range of transferability to high-density ALS data or even to more commonly available ALS data with densities of 10 points m$^{-2}$.

Based on the above sources of data variation, we tested five different scenarios that differed regarding the input data for model training, including the following combination of ULS, MLS data, and their sparsified versions:

- **Scenario 1**: Only ULS data (i.e. FOR-instance training data split)
- **Scenario 2**: Only MLS data
- **Scenario 3**: The combination of scenarios 1 and 2 was used as a baseline as a consistent dataset to understand the role of the data augmentation strategy (random sparsification) on the model's performance.
- **Scenario 4**: The combination of scenarios 1 and 2, plus their sparsified version at 1000 points m$^{-2}$, was used to provide an intermediate scenario between the lack of augmentation (scenario 3) and the fully augmented model (scenario 5).
- **Scenario 5**: The combination of scenarios 1 and 2, plus their sparsified versions at 1000, 500, 100, 75, 50, 25, and 10 points m$^{-2}$, represents the augmentation extreme where the training data was thinned to the level of ALS data.

## 3.2 Comparison of predictions at different resolutions

In order to evaluate the models' ability to transfer to ALS data at different resolutions, we compared the performance of the best-performing scenario on the FOR-instance dataset and the respective sparsified versions at all resolution steps. Through such analysis, we aimed to understand the extent to which the model can be transferred to ALS data with decreasing point densities without compromising performance. This analysis was limited to these data because the same analysis would not be meaningful for terrestrial datasets, which are by nature very dense and rarely come in the form of sparse data.

## 3.3 Performance across the canopy profile

We evaluated the best model's performance across the test dataset for different layers of the forest canopy. Given the lack of information for each tree on the canopy social status, similarly to the recent benchmark study by Cao et al. (2023) and by Xiang et al. (2023b), we used tree height to define the layers that each tree belonged to. For this purpose, we subdivided each test dataset into 5 m high vertical height bins and computed the metrics for each bin. Further, we visually assessed these results to better understand the relationship between the selected evaluation metrics and the model's predictions for the different layers.

## 3.4 Model

We chose the model architecture and implementation proposed by Xiang et al. (2023a). Point cloud segmentation network leverages a 3D convolutional neural network (CNN) as its core, enhanced by three parallel prediction heads. The first head handles semantic segmentation, assigning each point a class label, while the other two heads are focused on instance segmentation, identifying individual tree instances. Utilizing the Minkowski Engine library, our network strikes a balance between performance and computational efficiency. The semantic segmentation branch employs a multi-layer perception (MLP) to calculate class probabilities and distinguishes between tree and non-tree points. For instance segmentation, one branch predicts a 3D offset vector for tree centering, while another branch maps the points into a 5-dimensional embedding space to differentiate between tree instances. These embeddings are clustered using region-growing and mean-shift methods, respectively, and further refined by ScoreNet, a neural network that filters and merges the tree candidates based on ground truth overlap.

The model is trained end-to-end, utilizing a combined loss function comprising semantic, direction, regression, and score losses. During inference, NMS is performed on the clusters with the scores predicted by ScoreNet, leading to the final instance predictions (Jiang et al. 2020).

In the model's training process, Random Noise (Sigma: 0.01) and Random Rotation (Degrees: 180, Axis: 2) play critical roles in simulating real-world data variations. Introducing noise at a sigma of 0.01 adds slight perturbations to the data points, mimicking sensor inaccuracies or environmental disturbances typical in 3D scanning. This feature enhances the model's resilience against minor data inconsistencies. Concurrently, the implementation of random rotation, particularly along Z axis by up to 180 degrees, encompasses a broad spectrum of potential orientations, thereby ensuring the model's robustness to the orientation variance of objects, a factor that can significantly influence panoptic segmentation performance. Complementing these, Random Scale Anisotropic (Scales: [0.9, 1.1]) and Random Symmetry provide additional augmentation. These processes scale objects within a 90% to 110% range and apply asymmetrical reflection across specific axes, teaching the model to recognize objects amidst size fluctuations and various symmetrical orientations, prevalent scenarios in natural environments.

As previously mentioned, the model hyperparameters were constant throughout the tested scenarios.

### 3.5 Evaluation metrics

All the above analyses and comparisons were evaluated against the test datasets based on a point-wise matching of ground truth and predicted tree instance identifiers (see Figure 2). This consisted of matching predicted tree instances with ground truth based on the intersection over union (IoU) computed at the point level. According to this approach, we considered a tree correctly detected if a predicted tree instance had an IoU>0.5 with a ground truth instance. Based on the list of correctly detected trees, we developed tree-wise confusion matrices from which we obtained the counts for the true positive (TP), false positives (FP), and false negatives (FN) required to compute a selection of commonly used metrics for the evaluation of tree instance segmentation, which included:

$$Detection\ rate = \frac{Number\ of\ True\ Positives\ (TP)}{Total\ Number\ of\ Ground\ Truth\ Trees\ (GT)} \qquad \text{(Eq. 1)}$$

Where $TP$ represents trees that are correctly predicted (i.e. with IoU > 0.5) and $GT$ is the total number of ground truth trees.

$$Omission\ rate = \frac{Number\ of\ False\ Negatives\ (FN)}{Total\ Number\ of\ Ground\ Truth\ Trees\ (GT)} \qquad \text{(Eq. 2)}$$

Where, $FN$ represents ground truth trees that are not correctly predicted ($GT$ trees not in $TP$).

$$Commission\ rate = \frac{Number\ of\ False\ Positives\ (FP)}{Total\ Number\ of\ Predicted\ Trees\ (PT)} \quad \text{(Eq. 3)}$$

Where $FP$ represents the predicted trees that not correctly detected ($PT$ trees not in $TP$), and $PT$ represents the total number of predicted trees.

$$F1\ score = \frac{2xPrecisionxRecall}{Precision + Recall} \quad \text{(Eq. 4)}$$

Where:

$$Precision = \frac{TP}{TP + FP} \quad \text{(Eq. 5)}$$

$$Recall = \frac{TP}{TP + FN} \quad \text{(Eq. 6)}$$

In addition, following the approach by Wielgosz et al. (2023), we included the root mean square error (RMSE) for the tree height ($H$; m above ground). The RMSE was computed according to:

$$RMSE = \sqrt{\frac{1}{N}\sum_{i=1}^{N}(H_{GT,i} - H_{Pred,i})^2} \quad \text{(Eq. 7)}$$

Where $N$ is the total number of trees for which the metric is calculated, $H_{GT,i}$ is the height of the *i*-th tree in the ground truth, and $H_{pred,i}$ is the predicted height of the *i*-th tree.

In all cases, the tree height was computed as the difference between the tallest and shortest points for either the ground truth or the predicted tree instance. In addition to the inherent importance of tree height in determining other tree parameters (e.g., DBH or biomass), the inclusion of the tree height $RMSE$ is motivated by the fact that it represents a complementary measure to the detection rates and $F1\ score$, that captures the ability of the model to segment the whole length of the tree.

## 4. Results and discussion

### 4.1 Influence of data input selection and augmentation on performance

#### 4.1.1 Comparison between airborne and terrestrial data as source for model training

A first analysis of our results involved the comparison of scenarios 1 and 2 to understand the relative contribution of airborne (ULS) or terrestrial laser scanning (MLS) towards platform-agnostic models that can transfer across a broad range of test datasets. Results showed (see Figure 3 and Table 1) that, despite intrinsic differences in performances across the test datasets, the model trained using the ULS data had higher detection rates, lower commission errors, and higher F1-score than the MLS model, indicating its better ability to detect and segment individual trees. However, the difference between the ULS and MLS models was marginal for most of the tested datasets. However, a notable exception was observed in the TUWIEN dataset, where we found a substantial boost in the detection rates (i.e. approximately 20%) and a decrease in commission errors (reduction of approx. 20%) when using ULS data rather than the MLS data. This difference may be due to the complex forest structure (i.e. protected alluvial forest) in the TUWIEN dataset, which is present in the ULS but not captured in the MLS training data (i.e. capture only managed boreal forest).

Interestingly, as demonstrated by the ability of the ULS model to effectively work with high-resolution TLS datasets (like TreeLearn, Wytham woods, or LAUTx) and the MLS model's adaptability to airborne datasets (like FOR-instance) suggests that both terrestrial and airborne laser scanning data can train models transferable beyond their original platforms. This is significant as it implies the potential of using detailed, easily annotated terrestrial data (TLS or MLS) to train models that can also effectively handle sparser, more challenging airborne laser scanning (ALS) data, greatly broadening the use and scope of terrestrial data. Overall, these findings indicate the models' capability to identify general tree characteristics, offering a more robust tree segmentation approach compared to recently proposed methods such as TLS2trees (Wilkes et al. 2023) or Point2Tree (Wielgosz et al. 2023) that are by design non-agnostic as they rely on specific data characteristics of TLS and MLS data such as the visibility of stems in the lower parts of the canopy.

Further, Figure 3 also suggests that the magnitude of the difference between the performance of scenarios 1 and 2 was larger for the detection rate than the F1-score, meaning that once a tree was detected, the quality of the output segmentation was relatively consistent.

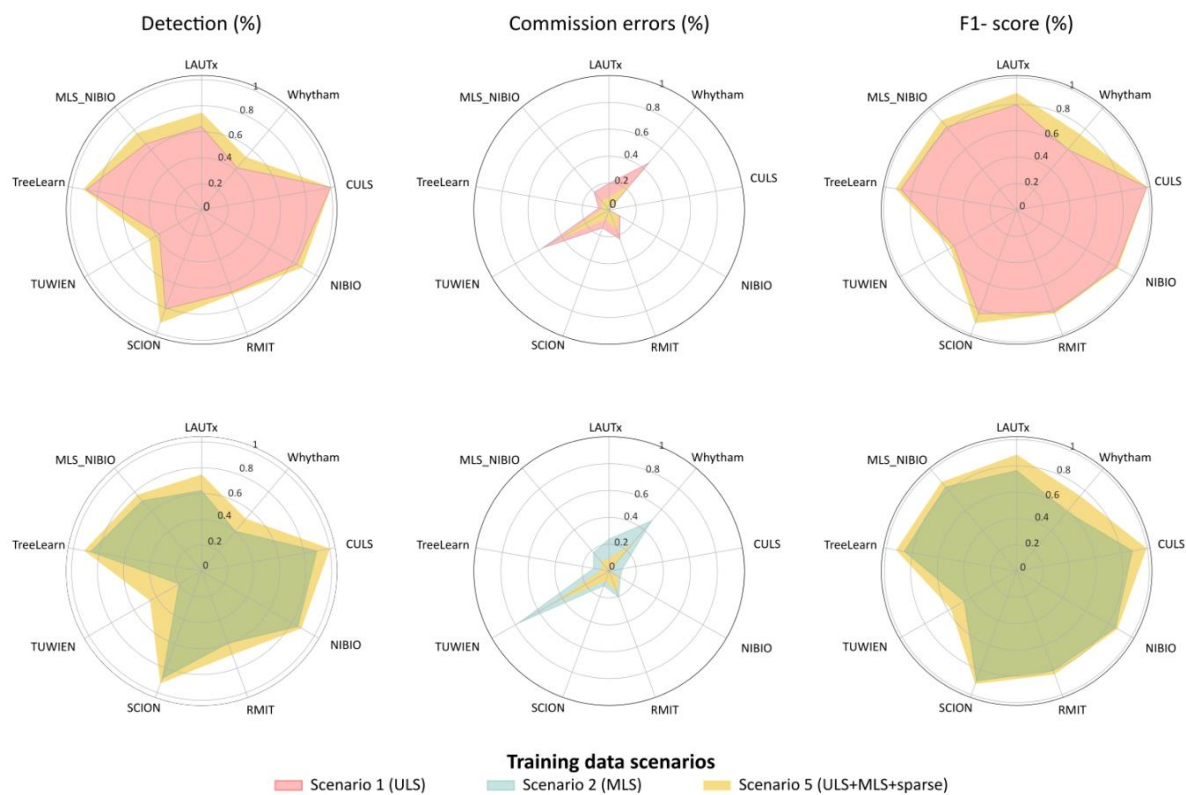

**Figure 3**. Radar charts comparing the performance of the model trained on using all data against the models using only the ULS (upper row) or the MLS data (lower row), for comparison we also included the performance of scenario 5 in yellow.

### 4.1.2 Comparison of augmentation strategies

In order to address the effect of the augmentation strategy on the performance of the segmentation on both ends of the scale in terms of point cloud density from detailed TLS data to airborne ULS data, we conducted a comparative analysis of scenarios 1 to 5, each representing a unique augmentation method. Our goal was to develop a model that could transfer to both ends of the scale.

The analysis, detailed in Table 1, underscores that scenarios 4 and 5 were consistently larger detection rates (i.e., on average 0.76 and 0.77) compared to scenarios 1 – 3 (on average 0.70, 0.67, and 0.72), meaning that overall, the augmentation through point cloud sparsification had a positive impact on the detection and segmentation performances. At the dataset level, the boost in detection rate was as large as 14% points for the TUWIEN dataset when going from scenario 3 (no augmentation) to scenario 5 (aggressive augmentation).

While both scenarios 4 and 5 showed similar performances, the latter stood out, particularly for its performance across most datasets. This highlights that an aggressive augmentation strategy, i.e. including point clouds as sparse as 10 pts m$^{-2}$, not only boosted the detection rates for airborne ULS datasets such as SCION (increase of 6% points) and TUWIEN (increase of 14.3% points) but also for very dense terrestrial datasets such as Wytham woods (increase of 10% points), LAUTx (increase of 5% points), or NIBO_MLS (increase of 6% points). The latter aspect is particularly interesting as it indicates that the model, trained on increasing abstract representations of tree point clouds, becomes adept at discerning and segmenting tree structures more effectively. A plausible explanation for this enhancement is the increase in the model's robustness thanks to the ability to abstract general tree patterns, compared to a more fragile model relying on specific data characteristics (i.e. stem visibility) determined by the platform and sensor type used to acquire the data. Such effect was most pronounced for datasets representing the most complex forest structures, such as Wytham woods or the TUWIEN datasets, underscoring the importance of including sparser representations when dealing with more complex forest 3D scenes.

**Table 1**. Summary of the performance metrics across the different studied scenarios and for each of the used test datasets.

| Scenario | Test dataset | Detection (%) | Omission (%) | Commission (%) | RMSE H (m) | F1 (%) |
|---|---|---|---|---|---|---|
| 1 (ULS) | Wytham woods | 0.42 | 0.58 | 0.46 | 5.11 | 0.60 |
| | TreeLearn | 0.89 | 0.11 | 0.08 | 6.31 | 0.89 |
| | LAUTx | 0.64 | 0.36 | 0.20 | 4.90 | 0.80 |
| | MLS_NIBIO | 0.66 | 0.34 | 0.17 | 3.34 | 0.82 |
| | CULS | 1.00 | 0.00 | 0.00 | 0.15 | 0.99 |
| | NIBIO | 0.84 | 0.16 | 0.09 | 3.37 | 0.87 |
| | TUWIEN | 0.37 | 0.63 | 0.57 | 4.43 | 0.54 |
| | SCION | 0.81 | 0.19 | 0.14 | 4.95 | 0.84 |
| | RMIT | 0.67 | 0.33 | 0.23 | 1.84 | 0.82 |
| 2 (MLS) | Wytham woods | 0.40 | 0.60 | 0.49 | 6.09 | 0.59 |
| | TreeLearn | 0.87 | 0.13 | 0.12 | 7.16 | 0.86 |
| | LAUTx | 0.62 | 0.38 | 0.24 | 5.99 | 0.77 |
| | MLS_NIBIO | 0.71 | 0.30 | 0.18 | **3.02** | 0.84 |
| | CULS | 0.90 | 0.10 | 0.10 | 4.94 | 0.89 |
| | NIBIO | 0.86 | 0.15 | 0.10 | 3.77 | 0.86 |
| | TUWIEN | 0.20 | 0.80 | 0.78 | 4.89 | 0.46 |
| | SCION | 0.88 | 0.12 | 0.12 | 2.40 | 0.89 |
| | RMIT | 0.59 | 0.41 | 0.21 | 1.74 | 0.81 |
| 3 (ULS + MLS) | Wytham woods | 0.43 | 0.57 | 0.43 | 4.81 | 0.62 |
| | TreeLearn | 0.86 | 0.14 | 0.09 | 5.99 | 0.87 |
| | LAUTx | 0.70 | 0.30 | 0.16 | 4.92 | 0.82 |
| | MLS_NIBIO | 0.71 | 0.30 | 0.17 | 3.63 | 0.82 |
| | CULS | 1.00 | 0.00 | 0.00 | 3.12 | 1.00 |
|  | NIBIO | 0.89 | 0.11 | 0.07 | 2.75 | 0.89 |
|  | TUWIEN | 0.31 | 0.69 | 0.65 | 4.58 | 0.51 |
|  | SCION | 0.86 | 0.14 | 0.11 | 3.28 | 0.88 |
|  | **RMIT** | **0.75** | **0.25** | **0.11** | 1.42 | **0.87** |
| 4 (ULS + MLS + sparse 1000) | Wytham woods | 0.48 | 0.52 | 0.34 | **3.88** | 0.67 |
|  | **TreeLearn** | **0.93** | **0.07** | **0.04** | 5.14 | **0.92** |
|  | LAUTx | 0.72 | 0.28 | 0.15 | 4.13 | 0.85 |
|  | MLS_NIBIO | 0.75 | 0.26 | 0.12 | 3.43 | 0.86 |
|  | CULS | 1.00 | 0.00 | 0.00 | 0.15 | 1.00 |
|  | **NIBIO** | **0.91** | **0.09** | **0.07** | 3.47 | **0.90** |
|  | TUWIEN | 0.43 | 0.57 | 0.48 | **3.05** | 0.58 |
|  | SCION | 0.90 | 0.10 | 0.10 | 2.89 | **0.93** |
|  | RMIT | 0.70 | 0.30 | 0.17 | 1.37 | 0.82 |
| 5 (ULS + MLS + sparse 1000, 500, 100, 10) | **Wytham woods** | **0.53** | **0.47** | **0.27** | 4.19 | **0.75** |
|  | TreeLearn | 0.92 | 0.09 | 0.05 | **0.09** | 0.92 |
|  | **LAUTx** | **0.75** | **0.25** | **0.10** | 3.11 | **0.89** |
|  | **MLS_NIBIO** | **0.77** | **0.24** | **0.09** | 3.44 | **0.88** |
|  | **CULS** | **1.00** | **0.00** | **0.00** | 0.15 | **1.00** |
|  | NIBIO | 0.88 | 0.12 | 0.08 | **3.41** | 0.88 |
|  | **TUWIEN** | **0.46** | **0.54** | **0.45** | 4.87 | **0.57** |
|  | **SCION** | **0.92** | **0.08** | **0.08** | 1.83 | **0.91** |
|  | RMIT | 0.69 | 0.31 | 0.17 | **1.29** | 0.84 |

Data augmentation is widely recognized as a valuable technique for enhancing the robustness of models, particularly in scenarios where training data are abundant, such as labelled 3D forest point clouds. From a machine learning standpoint, data augmentation contributes significantly to improved generalization. This is especially true in cases where sparse convolutions are utilized. Exposing the model to varying levels of sparsity allows it to generate more effective features for datasets where the relationships between points are less immediate and more dispersed.

Moreover, it is important to highlight that employing convolutional kernels with larger aspect ratios could further optimize the model's performance, leveraging the benefits of sparsity-based augmentation. This approach could enable the model to better capture and interpret the nuanced spatial relationships in the data, leading to more accurate and reliable predictions.

However, for the sake of maintaining consistency and comparability across different iterations of the model training, the authors have chosen not to alter the model's core structure or introduce these modifications at this stage. This decision ensures that any improvements or changes in performance can be directly attributed to the data augmentation techniques employed rather than to alterations in the model's architecture. This approach not only preserves the integrity of the comparative analysis but also opens avenues for future research to explore the potential benefits of these architectural modifications in enhancing model performance.

Regarding point cloud augmentation for tree instance segmentation tasks, Xiang et al. (2023b) explored a range of augmentation strategies from common methods, such as noise, rotation, scaling, reflection, dropout, and elastic deformation, to the more sophisticated TreeMix approach. As in our study, Xiang et al. (2023b) found a positive effect of the augmentation on the detection rates and segmentation quality. In this context, our study provided a more naive and simple approach to augmentation. Despite that, we found a similar (5.3% points) improvement in performance from scenario 1 to scenario 5 compared to the increase in detection rate reported by (Xiang et al. 2023b) between their basic and most advanced augmentation scenario (4% points). Thus, where comprehensive labelled training datasets for point cloud tree segmentation are scarce, data augmentation holds an important role. While sophisticated point cloud augmentation methodologies currently allow for the simulation of new forest 3D scenes from existing point clouds such as the

TreeMix approach proposed by Xiang et al. (2023b) or the Helios++ simulation software (Winiwarter et al. 2022), they do come with increased computational demands. Our findings suggest that even elementary augmentation strategies can yield notable performance improvements. Nevertheless, future studies should explore the cumulative effects of combining various augmentation techniques to identify an optimal strategy that maximizes the benefits of available methods while ensuring computing efficiency at the same time.

### 4.2 Performance for different point cloud resolutions

When looking at the performance of the best model (i.e. scenario 5) on the FOR-instance point clouds sparsified at different levels of point cloud density (see Figure 4 and Table A.1), it is possible to immediately notice that the model performance was stable for all metrics in point clouds with point densities > 50 points $m^{-2}$. However, the performance dropped when predicting on the sparsest point clouds (10 points $m^{-2}$) mainly due to a substantial increase in omission rates (i.e. approximately 20% points). The commission errors and quality of the segmentation (F1 score) were less affected by the decrease in point cloud density.

Such a result indicates that the proposed methods might be optimally suitable for dense ALS data captured from low-flying aircrafts or helicopters. While still limited in operational settings, these data types are increasingly being used by researchers and the forest industry in Nordic countries (e.g. Hakula et al. 2023; Hyyppä et al. 2022; Persson et al. 2022). Further potential use of the proposed method is for the segmentation of drone laser scanning data, including both data from consumer-grade laser scanning data (e.g. DJI Zenmuse L1 or Velodyne VLP16) as well as survey-grade laser scanning data (e.g. Riegl VUX-1UAV or miniVUX). To provide a visual understanding of the performance of our model on real data, Figure A.1 (see Appendix) shows the output segmented tree instances on dense airborne laser scanning data captured either from a helicopter (920 points $m^{-2}$), manned aircraft (665 points $m^{-2}$), and consumer-grade drone laser scanning data (661 points $m^{-2}$). Given that these datasets have not been manually annotated or have field-based ground data, it is only possible to do any visual assessment of the segmentation that shows promising results across all three datasets.

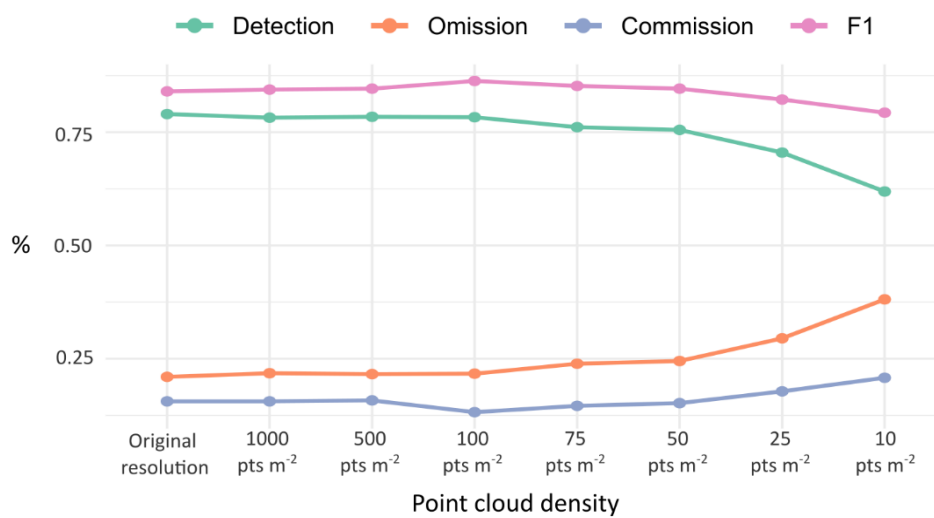

**Figure 4**. Average performance across the FOR-instance dataset according to the evaluation metrics and across the increasing sparsification steps tested in this study.

When comparing our results on the sparsified versions of the FOR-instance dataset with those by Straker et al. (2023), who using a similar random sparsification strategy benchmarked a YOLOv5 canopy height model (i.e. 2D raster) segmentation, we can see that, although Straker et al. (2023) performed the analysis on a previous beta version of FOR-instance which included an additional dataset, our model performed consistently better than the method by Straker et al. (2023) in terms of detection rates for the original resolution (79.1 versus 60.7), and the sparsified versions at 500 points m$^{-2}$ (0.78 versus 0.64) and 10 points m$^{-2}$.

### 4.3 Performance across canopy profile

The analysis of the performance of the model trained on scenario five across different canopy layers (see Figure 5) revealed that the performance varied depending on the forest type (deciduous versus coniferous forests and complex multilayered forests versus single layered forests) and for specific canopy layers (well-separated layers versus vertically connected layers), but there was no clear pattern indicating poorer performances for specific input data sources, meaning that the model was transferable to terrestrial as well as to dense airborne laser scanning data. While the available data did not provide information on the specific social distribution of trees, by using tree height as a proxy, we found (see Figure 5) that the model was able to segment trees throughout the canopy vertical profile, including understory suppressed trees (trees shorter than 10 m; detection rates in the range 0.2 - 0.8), co-dominant, and dominant trees.

The performance was generally poorer for multilayered broadleaved forests compared to coniferous forests. Part of the reason might relate to the fact that the largest proportion of the training data was from managed boreal forests, and thus, improvements to our model could be obtained by expanding the training data to include a broader range of forest types. A particular case was that of the Wytham Woods dataset, which, composed of approximately 53% of the trees shorter than 15 m, is characterized by a dense understory layer composed of young saplings and multi-stem trees. The poor detectability of these trees was likely the driver of the overall poor performance of our method on the Wytham data. Furthermore, it is important to remember that the Wytham Woods dataset was collected using a survey-grade TLS scanner, which was not represented in the training data.

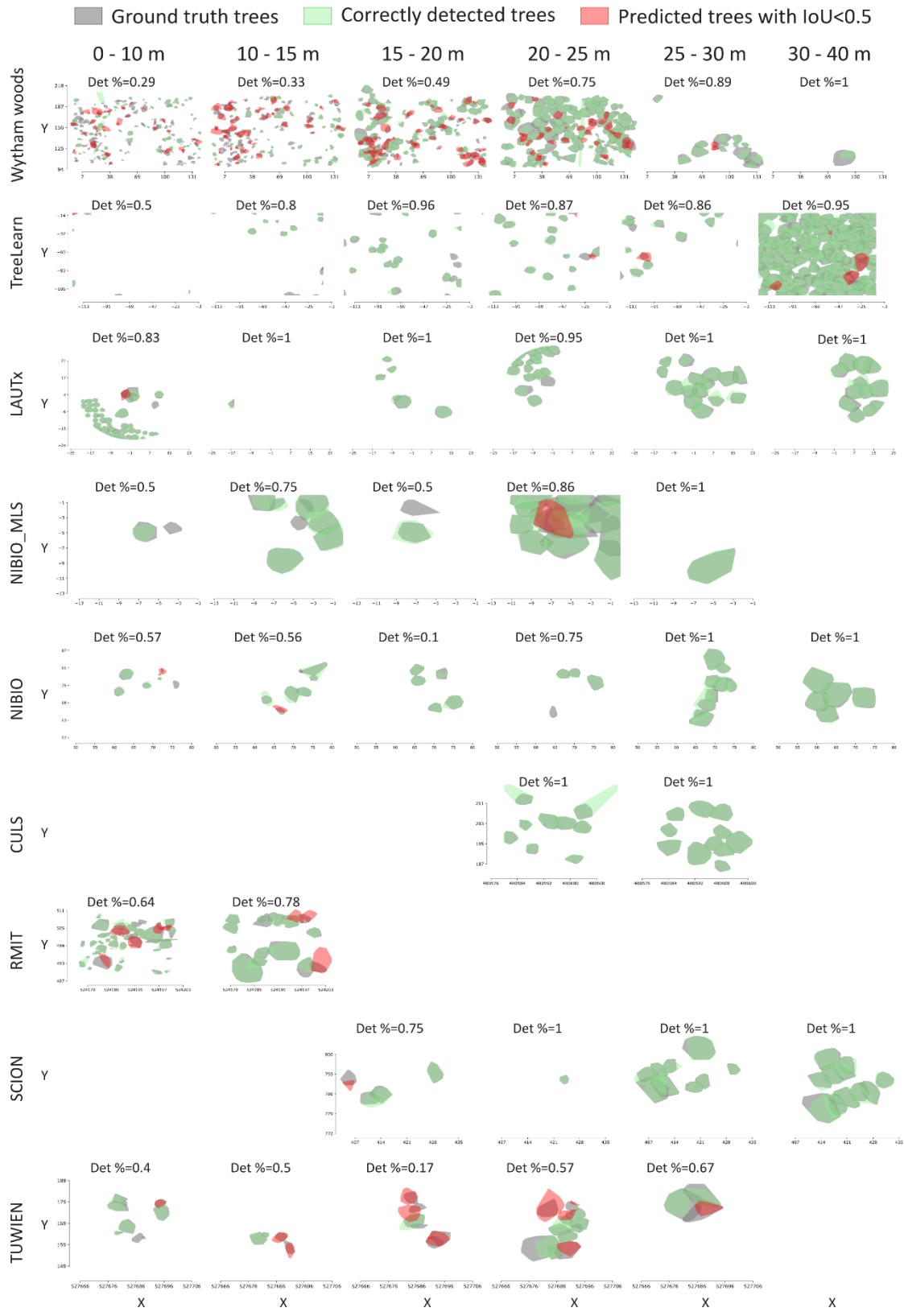

**Figure 5**. Visual comparison of the performance in terms of detection rate (%) for different layers of the canopy (5 m high bins). For ensuring clarity only the detection rate for each of the slices is reported.

## 4.4 Computational efficiency

In the evaluation of computational efficiency, we propose "Data Processed per Core per Unit Time" to assess the performance of processing systems. This metric quantifies the efficiency with which a system processes data. To compute this metric, three essential elements are considered: the data size processed, measured in megabytes (MB); the number of cores, which indicates the computational resources employed; and the processing time, standardized to minutes for consistency. The proposed computational efficiency ($CE$) metric was computed as the ratio between the amount of processed data (in MB) and the product of the number of cores and processing time.

$$CE = \frac{Data\ processed\ (MB)}{Number\ of\ Cores * Processing\ time\ (minutes)} \quad \text{(Eq. 8)}$$

By employing this metric, one can compare the processing efficiencies of various systems and configurations. For TLS2trees, we utilized the values reported by Wilkes et al. (2023). Table 2 illustrates that in terms of $CE$ the method presented in this paper performs an order of magnitude better than that reported for TLS2Trees (Wilkes et al. 2023). It is worth noting that we assume that a single GPU is used for the two approaches and 64 cores for our approach, and 200 cores for graph-based approach as given in the paper (Wilkes et al. 2023). The assumption is that both solutions use a single GPU without accounting for very details of gpu architectures.

**Table 2**. Computational efficiency of the proposed method against TLS2trees.

| Method | Tested dataset | Time (min) | Size (MB) | $CE$ (MB/Core/Min) |
|---|---|---|---|---|
| This study | LAUTx | 403.12 | 11264 | 0.44 |
|  | Wytham woods | 309.51 | 5120 | 0.26 |
|  | FOR-instance | 49.13 | 1638 | 0.52 |
|  | TreeLearn | 57.88 | 480 | 0.13 |
|  | NIBIO_MLS | 48.92 | 1331 | 0.42 |
|  | **average** | **173.712** | **3966.6** | **0.354** |
| TLS2trees | RUSH | 101 | 300 | 0.015 |
|  | NOU | 1245 | 8212.5 | 0.033 |
|  | MLA | 1370 | 13487.5 | 0.05 |
|  | **average** | **905.33** | **7333.33** | **0.032** |

In our system performance evaluation, it is crucial to acknowledge the diversity of processing units employed, encompassing both Central Processing Units (CPUs) and Graphics Processing Units (GPUs). Our machine utilizes an Intel® Xeon® Gold 6246R CPU, operating at 3.40GHz with a cache size of 36608 KB. This CPU, belonging to the Xeon family, is known for its robust performance in enterprise and data center environments, offering high processing power and efficiency. The detailed specifications of the CPU, such as its 64 cores, advanced vector extensions (AVX-512), and other architectural features, play a pivotal role in determining the system's computational capabilities and efficiency, as measured by the $CE$.

Additionally, our system is equipped with a NVIDIA GRID V100S-16Q GPU, a powerful graphics unit designed for demanding computational tasks. The GPU's specifications, including its memory usage and CUDA version, are integral to understanding the system's overall performance, especially in tasks that are parallelizable and can leverage the GPU's architecture.

## 4.5 Visual assessment of predictions

A general visual assessment of the output point clouds from the model trained in scenario five can be found in Figure 6. Here, we can see the relatively high segmentation quality for most datasets. However, as found in the previous section, it is also clear that the performance on the most complex forest types (i.e. Wytham woods and TUWIEN) was affected by the lack of detection of some understory trees or co-dominant trees with intertwined crowns.

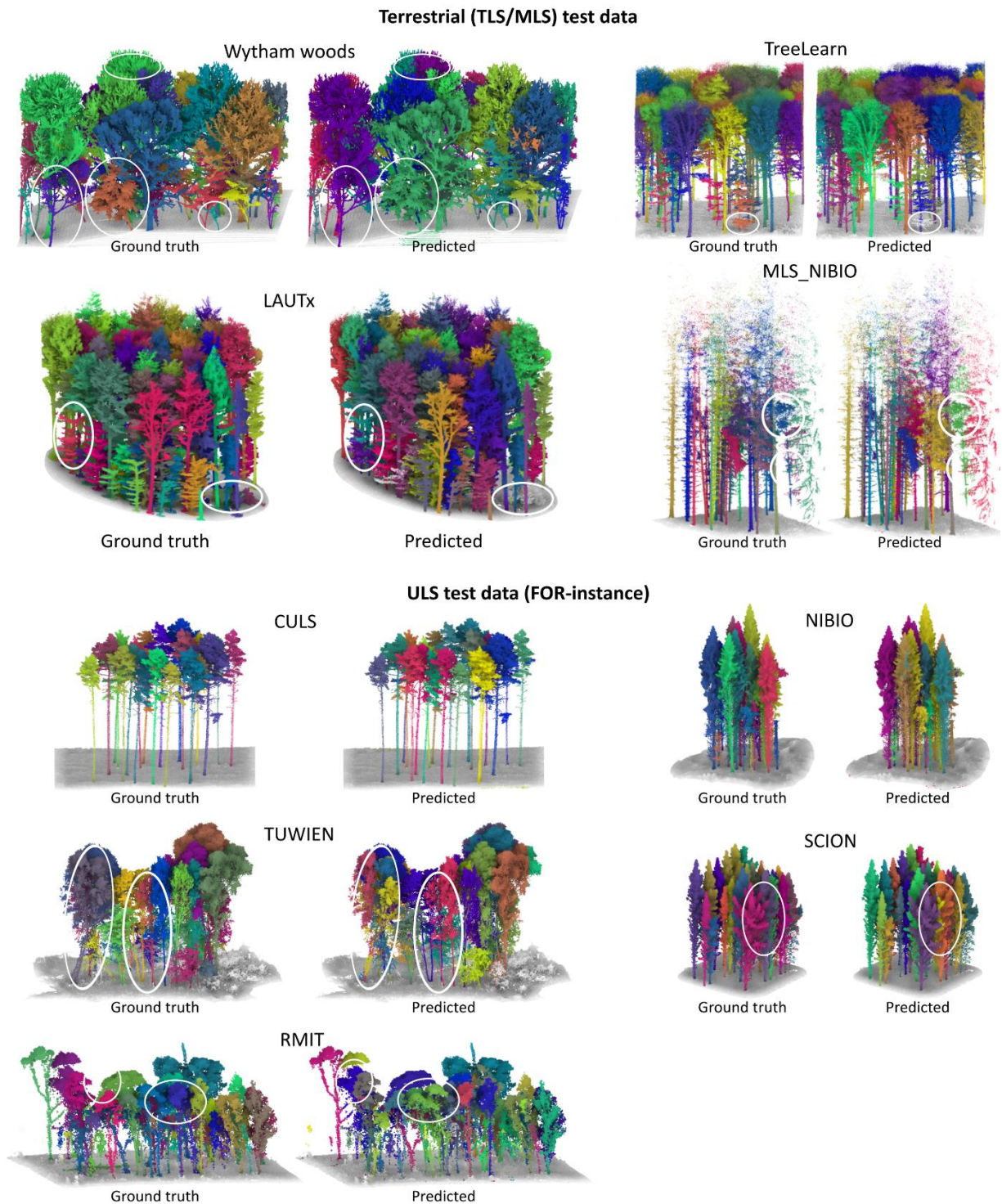

**Figure 6**. Example point clouds from the test dataset and the respective predictions using scenario 5. The white ellipses highlight some of the main issues in the output segmented instances.

## 4.6 Advancing the state-of-the-art

In order to assess the performance of our proposed method to the current SOTA we benchmarked our results against those of previous studies relying on the same benchmark datasets. Benchmarking against public datasets represents the most objective and efficient way to track the progress in model development (Lines et al. 2022). However, it is only until very recently that ML-ready dataset for point cloud tree instance segmentation have been publicly available. In this context our study represents an important step forward towards advancing our ability to track model development. In specific, the comparison with previous studies done on the used benchmark test datasets (see Table 3), showed the following regarding the fully independent datasets:

- **Whytham woods dataset**: While there exist previous studies addressing tree segmentation of TLS data in Wytham woods (e.g. Wilkes et al. 2023; Xu et al. 2023), these have been done on a sub-area (1 ha) of the 1.4 hectares area published by Calders et al. (2022) and thus are not comparable to our results. Thus, our study provides a first baseline performance for this test dataset that is particularly valuable thanks to the complexity of the forest structure. The performance found for this dataset (detection: 0.53; omission: 0.47; commission: 0.27; RMSE H: 4.2; and F1-score: 0.88) was amongst the poorest across the different datasets. A similar conclusion was drawn by Wilkes et al. (2023), who found this to be most challenging dataset for tree segmentation in their study. While the inherent complexity of the forest structure in Wytham woods makes the task of segmenting individual trees challenging, improvements to our method could derive from training the model using also labelled survey-grade TLS data from complex structures found for example in tropical forests.
- **TreeLearn**: The comparison with the study by Henrich et al. (2023) shows that, in terms of F1-score our method (0.92) performed worse than theirs (0.98), which was on the other hand trained using similar data from similar areas and thus likely inflated.
  Our model showed the highest detection rates across the tested datasets and considering that this was truly an independent dataset from the one used for training the model, we can conclude that the developed model, is suitable for prediction on MLS data collected in relatively simple forest structures characterized by a single main canopy layer and with sporadic and vertically separated individual understory trees. As this dataset was fully independent from the training set the metrics can be considered as absolute values, hence indicating the true performance against these data.
- **LAUTx**: the comparison with baseline values defined by Wielgosz et al. (2023) for Point2Tree and TLS2trees on the LAUTx dataset, showed that our method outperformed Point2Tree and TLS2trees by approximately 20% points in improvements in terms of detection, omission, commission, and F1 score. The improvement was even larger (approximately 25-30%) when comparing against the TLS2trees performance reported by Wielgosz et al. (2023) on the LAUTx dataset. Hence, the proposed method represents a substantial leap in the SOTA over the LAUTx dataset compared to both Point2Tree and TLS2trees. Given that this dataset, like also the TreeLearn and the NIBIO_MLS data, were collected using the same MLS scanner (i.e. Geoslam Horizon), the poorer performance compared to the TreeLearn dataset is likely attributable primarily to the increased complexity of the forest structure in the LAUTx dataset, where some of the plots (e.g. see example LAUTx plots in Figure 6) were characterized by interconnected multiple layers. Similarly to the Wytham woods and TreeLearn, the LAUTx dataset was fully independent from the training data and thus the obtained metrics values can be considered as absolute values representing the true performance of our method on new data.

Differently to the above datasets the NIBIO_MLS datasets and the FOR-instance test datasets were split from the training data but contiguous in space with the data used for training the models and as such it does not represent a fully independent evaluation. Therefore, for the below cases, the metric values should be taken as relative values, i.e. for comparison of methods on the same data, rather than absolute values indicating the general performance of our method across the tested forest types. In specific we found the following for each of the datasets:

- **NIBIO_MLS**: consistently with the comparison of the SOTA for the LAUTx dataset, when comparing with the performance of Point2Tree and TLS2trees, we found a substantial improvement of the SOTA corresponding to a 17 – 20% boost in detection rates, and a 26 – 27% boost in the F1-score (i.e. segmentation quality). Such degree of improvement on these key segmentation metrics shows the clear benefit of utilizing our proposed approach over the previously proposed Point2Tree and TLS2trees on mobile laser scanning point clouds. On the other hand, the decrease in commission errors (i.e. 2 – 5%) and RMSE of the tree height (i.e.7 – 20 cm) was more modest, indicating that either of the methods provides low commission errors and relatively accurate tree heights.
- **FOR-instance (CULS)**: For this dataset the comparison with baseline metrics showed no improvement as all of the previous studies obtained a nearly perfect segmentation due to the simple structure of the forest (i.e. single layered open Scots pine forest).
- **FOR-instance (NIBIO)**: our results on the NIBIO dataset showed that our model's performance closely mirrored that of the model developed by Xiang et al. (2023b). Our approach, which utilized a basic binary semantic segmentation combined with straightforward augmentation techniques, surprisingly matched the effectiveness of the method by Xiang et al. (2023b) which was more complex and included advanced augmentation methods. This result underscores the potential of simpler strategies like random point sparsification in achieving results comparable to those obtained with more sophisticated techniques.
- **FOR-instance (TUWIEN)**: for this dataset we found that our model significantly underperformed compared to the SOTA model by Xiang et al. (2023b). This gap in performance suggests the critical importance of the enhancements they implemented, such as the incorporation of multiple semantic classes (ground, tree stems, branches, and leaves) and the use of synthetic forest point cloud scenes for augmentation. In particular, in their method, the availability of information on the location of the stems might provide a more robust means of identifying tree instances.
- **FOR-instance (SCION)**: As for the results on the NIBIO dataset, the ones on the SCION data revealed very similar performance between our proposed model and the one by Xiang et al. (2023a).
- **FOR-instance (RMIT)**: While the results showed that our method had larger detection rates, lower commission errors, and largest segmentation accuracy (i.e F1-score) than the model by Xiang et al. (2023a) the improvement was marginal.

**Table 3**. Summary of the performance metrics for existing studies against the our best results for any of the scenarios on the tested datasets. The bold font indicates state-of-the-art performance on each of the evaluated metrics.

| Test dataset | Method | Detection (%) | Omission (%) | Commission (%) | RMSE H (m) | F1-score (%) |
|---|---|---|---|---|---|---|
| Wytham woods | Our method | **0.53** | **0.47** | **0.27** | **4.2** | **0.88** |
| TreeLearn | TreeLearn (Henrich et al. 2023) | - | - | - | | 0.98 |
| | Our method | **0.93** | **0.07** | **0.03** | 5.1 | 0.92 |
| LAUTx | Point2tree (Wielgosz et al. 2023) | 0.55 | 0.45 | 0.15 | **2.7** | 0.67* |
| | TLS2trees (Wielgosz et al. 2023) | 0.405 | 0.6 | 0.3 | 3.6 | 0.63* |
| | **Our method** | **0.75** | **0.25** | **0.09** | 3.1 | **0.88*** |
| NIBIO MLS | Point2tree (Wielgosz et al. 2023) | 0.57 | 0.43 | **0.07** | 3.47 | 0.61* |
| | TLS2trees (Wielgosz et al. 2023) | 0.59 | 0.41 | 0.14 | 3.6 | 0.62* |
| | **Our method** | **0.77** | **0.23** | 0.09 | **3.4** | **0.88*** |

| Dataset | Reference | | | | | |
|---|---|---|---|---|---|---|
| NIBIO | Straker et al. (2023) | 0.67 | 0.33 | - | | - |
| | Xiang et al. (2023b) | **0.88** | **0.12** | **0.03** | | **0.92**\*\* |
| | Our method | 0.88 | 0.12 | 0.09 | 3.4 | 0.88\* |
| CULS | Straker et al. (2023) | 1 | 0 | - | | - |
| | Xiang et al. (2023b) | 1 | 0 | 0.13 | | 0.93\*\* |
| | **Our method** | **1** | **0** | **0** | **0.14** | **0.99** |
| SCION | Straker et al. (2023) | 0.86 | 0.14 | - | | - |
| | Xiang et al. (2023b) | 0.87 | 0.13 | **0.04** | | **0.91**\*\* |
| | **Our method** | **0.92** | **0.08** | 0.07 | **1.7** | 0.91 |
| RMIT | Straker et al. (2023) | 0.58 | 0.42 | - | | - |
| | Xiang et al. (2023b) | 0.64 | 0.36 | 0.24 | | 0.7\*\* |
| | **Our method** | **0.69** | **0.31** | **0.17** | **1.3** | **0.83** |
| TUWIEN | Straker et al. (2023) | 0.2 | 0.8 | - | | - |
| | Xiang et al. (2023b) | **0.71** | **0.29** | **0.32** | | **0.69**\*\* |
| | Our method | 0.46 | 0.54 | 0.45 | 4.8 | 0.57 |

\* local computation of metric (i.e. F1-score is computed only using correctly detected trees)
\*\* global computation of metric (i.e. F1-score is computed for all trees)

Based on the above, we found that our method places itself at the forefront of the SOTA for instance segmentation of dense laser scanning point clouds (i.e. TLS, MLS, ULS) based on the following aspects:

- We defined independent baseline metrics for tree instance segmentation of prominent datasets such as the Wytham woods and TreeLearn datasets.
- Performed similar to, or even better than the baseline held and by Xiang et al. (2023a) for the FOR-instance dataset
- Performed substantially better than the baseline defined by the Wielgosz et al. (2023) for the LAUTx and NIBIO_MLS datasets.

When looking more closely at the performances on the overall FOR-instance dataset and its sparsified versions (see Table 4) we found that, despite the baseline defined by Straker et al. (2023) was computed using a previous version of FOR-instance (including an additional NIBIO_2 dataset), our results showed an overall boost of the detection rates of approximately 20% points for the full resolution, 14% points for the sparsified version at 500 points $m^{-2}$, and 8.2% points for the sparsified version at 10 points $m^{-2}$. While the magnitude of the improvement was substantial, such findings are not particularly surprising since the method by Straker et al. (2023) relied on the rasterized canopy height model and thus inherently eliminating the segmentation of dominated trees. Overall, the benchmarking of the performance on the sparsified version of the FOR-instance data showed that our method substantially advances the SOTA for ALS-like data.

**Table 4**. Summary of the state-of-the-art on the aggregated FOR-instance dataset and its sparsified version that are comparable with the study by Straker et al. (2023). The bold font indicates state-of-the-art performance on each of the evaluated metrics.

| Test dataset | Reference | Detection (%) | Omission (%) | Commission (%) | RMSE H (m) | F1-score (%) |
|---|---|---|---|---|---|---|
| FOR-instance | Straker et al. (2023) | 60.7 | 39.3 | - | - | |
| | Xiang et al. (2023a) | - | - | - | - | 68.9 |
| | Our study | **79.1** | **20.9** | 15.5 | 2.3 | 83.9 |
| FOR-instance 1000 | Our study | **78.3** | **21.7** | 15.6 | 4.1 | 84.5 |
| FOR-instance 500 | Straker et al. (2023) | 64.2 | 35.8 | - | - | - |
| | Our study | **78.5** | **21.5** | 15.7 | 2.6 | 84.5 |
| FOR-instance 10 | Straker et al. (2023) | 50.5 | 49.5 | | | |
| | Our study | **58.7** | **41.3** | 21.3 | 5.8 | 77.7 |

## 4. Conclusion

The study illustrates that it is feasible to train fully agnostic models that can be applied to the full spectrum of resolutions available of laser scanning data both from airborne and terrestrial platforms. The improved model performance when including different types of datasets (platforms and densities) as well as augmentation (specification) was large and shows a very promising avenue for model development for forest lidar point clouds. The results illustrate that we may be moving towards an alignment of approaches for segmentation for terrestrial and airborne point clouds which in the forest remote sensing community traditionally been two separate fields with separate segmentation strategies.

The study shows the strength of utilizing available open-source benchmarking datasets both for model development as well as for performance assessment. Compared to the tradition in the forest remote sensing community, where each study normally has a separate dataset from a given region, this provides a new way of working and a transparent way forward to understand improvements in the SOTA of the many new algorithms that undoubtedly will be presented in the coming years with overall improvements in AI algorithms.

SegmentAnyTree shows a notable leap in SOTA in particular thanks to the improved detection and segmentation of dominated trees under the canopy. Our method furthers the SOTA also in terms of reducing computing resources compared to previous methods. At the same time, the presented results provide an important benchmark for some of the most prominent existing open-source datasets.

SegmentAnyTree can clearly be further improved by providing the training data including data from survey grade TLS and more complex forest types. However, for conniforous-dominated forest the performance is already very good.

## Aknowledgements

## Authors contributions

MW and SP contributed equally to this manuscript, RA jointly contributed to the conceptualization of the paper and contributed with input on the manuscript, BX and KS contributed with comments on the manuscript.

## Data availability

The data used in this study is publicly available at the following links:

- Wytham woods dataset: https://zenodo.org/records/7307956
- TreeLearn dataset: https://data.goettingen-research-online.de/dataset.xhtml?persistentId=doi:10.25625/VPMPID
- FOR-instance dataset: https://zenodo.org/records/8287792
- LAUTx dataset: https://zenodo.org/records/6560112
- NIBIO_MLS dataset: will be publish upon acceptance of the manuscript

The code for training and prediction using the proposed deep learning model will be published upon acceptance.

## Declaration of AI and AI-assisted technologies in the writing process

During the preparation of this work the authors used OpenAI's ChatGPT v4.0 in order to improve readability. After using using this tool the authors reviewed and edited the content as needed and take full responsibility for the content of the publication.

# APPENDIX

**Table A.1**. Performance of the model trained under the scenario 5 across the different FOR-instance sparsified datasets.

| Sparsified datasets | Test dataset | Detection (%) | Omission (%) | Commission (%) | RMSE H (m) | F1 (%) |
|---|---|---|---|---|---|---|
| Original resolution | CULS | 1.00 | 0.00 | 0.00 | 0.15 | 1.00 |
| | NIBIO | 0.88 | 0.12 | 0.08 | 3.41 | 0.88 |
| | TUWIEN | 0.46 | 0.54 | 0.45 | 4.87 | 0.57 |
| | SCION | 0.92 | 0.08 | 0.08 | 1.83 | 0.91 |
| | RMIT | 0.69 | 0.31 | 0.17 | 1.29 | 0.84 |
| Sparsified 1000 pts m$^{-2}$ | CULS | 1.00 | 0.00 | 0.00 | 5.72 | 0.98 |
| | NIBIO | 0.87 | 0.13 | 0.07 | 3.36 | 0.90 |
| | TUWIEN | 0.40 | 0.60 | 0.48 | 5.82 | 0.59 |
| | SCION | 0.92 | 0.08 | 0.08 | 4.04 | 0.91 |
| | RMIT | 0.72 | 0.28 | 0.15 | 1.52 | 0.84 |
| Sparsified 500 pts m$^{-2}$ | CULS | 1.00 | 0.00 | 0.00 | 0.12 | 1.00 |
| | NIBIO | 0.90 | 0.10 | 0.05 | 2.72 | 0.92 |
| | TUWIEN | 0.43 | 0.57 | 0.44 | 4.75 | 0.58 |
| | SCION | 0.90 | 0.10 | 0.10 | 3.85 | 0.91 |
| | RMIT | 0.69 | 0.31 | 0.20 | 1.59 | 0.82 |
| Sparsified 100 pts m$^{-2}$ | CULS | 1.000 | 0.000 | 0.00 | 0.37 | 0.99 |
| | NIBIO | 0.84 | 0.16 | 0.08 | 2.92 | 0.89 |
| | TUWIEN | 0.58 | 0.42 | 0.29 | 4.23 | 0.71 |
| | SCION | 0.88 | 0.12 | 0.07 | 2.22 | 0.91 |
| | RMIT | 0.61 | 0.39 | 0.22 | 1.82 | 0.80 |
| Sparsified 75 pts m$^{-2}$ | CULS | 1.00 | 0.00 | 0.00 | 1.27 | 1.00 |
| | NIBIO | 0.80 | 0.20 | 0.11 | 3.21 | 0.87 |
| | TUWIEN | 0.56 | 0.44 | 0.31 | 4.11 | 0.70 |
| | SCION | 0.88 | 0.12 | 0.06 | 2.93 | 0.92 |
| | RMIT | 0.56 | 0.44 | 0.25 | 1.87 | 0.78 |
| Sparsified 50 pts m$^{-2}$ | CULS | 1.00 | 0.00 | 0.00 | 2.96 | 1.00 |
| | NIBIO | 0.79 | 0.22 | 0.11 | 3.48 | 0.87 |
| | TUWIEN | 0.55 | 0.45 | 0.33 | 4.28 | 0.69 |
| | SCION | 0.86 | 0.14 | 0.08 | 3.79 | 0.90 |
| | RMIT | 0.57 | 0.43 | 0.24 | 1.99 | 0.78 |
| Sparsified 25 pts m$^{-2}$ | CULS | 0.98 | 0.03 | 0.02 | 4.11 | 0.97 |
| | NIBIO | 0.71 | 0.29 | 0.14 | 4.48 | 0.84 |
| | TUWIEN | 0.54 | 0.46 | 0.32 | 5.61 | 0.69 |
| | SCION | 0.85 | 0.15 | 0.10 | 5.18 | 0.88 |
| | RMIT | 0.45 | 0.56 | 0.31 | 2.29 | 0.73 |
| Sparsified 10 pts m$^{-2}$ | CULS | 0.89 | 0.11 | 0.06 | 7.95 | 0.92 |
| | NIBIO | 0.62 | 0.39 | 0.13 | 5.07 | 0.83 |
| | TUWIEN | 0.46 | 0.54 | 0.35 | 5.48 | 0.67 |
| | SCION | 0.81 | 0.19 | 0.13 | 5.34 | 0.85 |
| | RMIT | 0.32 | 0.68 | 0.37 | 2.40 | 0.69 |

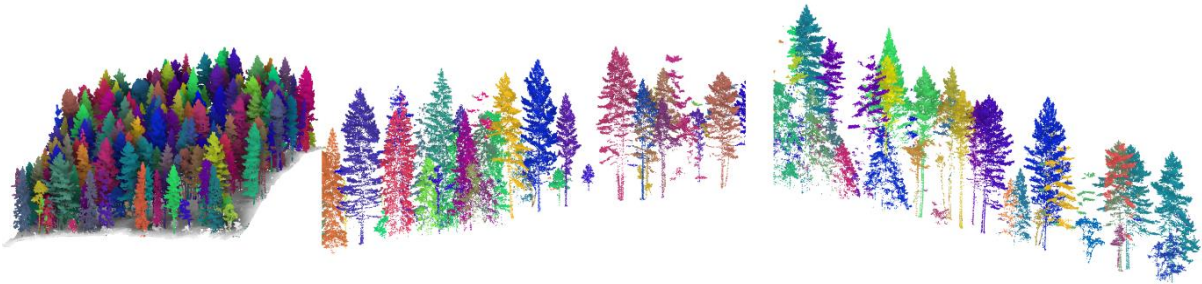

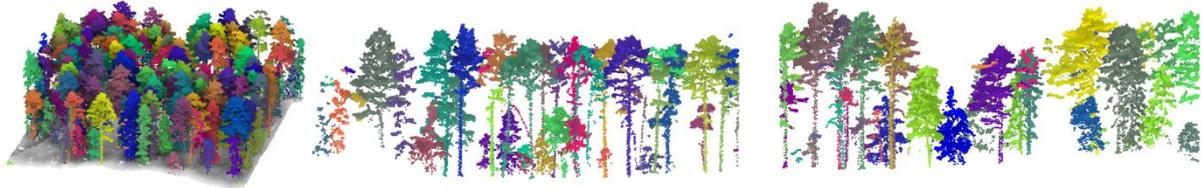

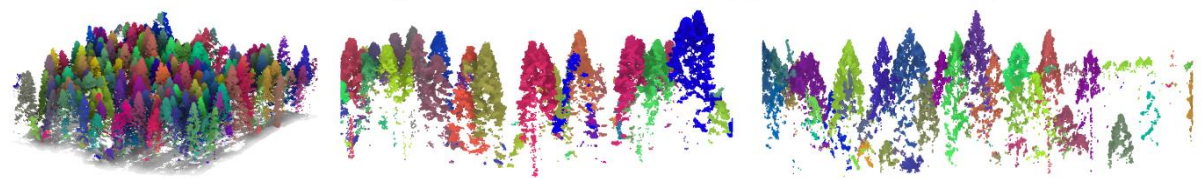

**Figure A.1**. Examples of predictions using the model trained in scenario 5 (i.e. full augmentation) on data unseen by the model including dense airborne laser scanning point clouds captured either with survey grade scanners mounted either helicopters, or on manned aircraft, or using consumer grade drone laser scanning data. In specific, the figure highlights the main improvement compared to previous methods applied to ALS data, i.e. the ability of the model of segmenting understory or dominated trees even when they are found in dense groups.